\documentclass[letterpaper]{article} 
\usepackage{aaai2026}  
\usepackage{times}  
\usepackage{helvet}  
\usepackage{courier}  
\usepackage[hyphens]{url}  
\usepackage{graphicx} 
\usepackage{natbib}  
\usepackage{caption} 
\usepackage{newfloat}
\usepackage{listings}
\DeclareCaptionStyle{ruled}{labelfont=normalfont,labelsep=colon,strut=off} 
\usepackage{hyperref}
\usepackage{amsmath}
\usepackage[utf8]{inputenc} 
\usepackage[T1]{fontenc}    
\usepackage{url}            
\usepackage{booktabs}       
\usepackage{amsfonts}       
\usepackage{nicefrac}       
\usepackage{microtype}      
\usepackage{xcolor}         
\usepackage{graphicx}       
\usepackage{placeins}          
\usepackage{todonotes} 
\usepackage{subcaption}
\usepackage{algorithm}
\usepackage[noend]{algpseudocode}
\usepackage[shortlabels]{enumitem}
\usepackage[capitalize,noabbrev]{cleveref}

\newcommand{\R}{\mathbb{R}}
\newcommand{\RT}{MRE}  
\DeclareMathOperator{\EX}{\mathbb{E}}  
\DeclareMathOperator{\Var}{\mathrm{Var}}  

\title{Mixture of Raytraced Experts}

\author {
    \equalcontrib Andrea Perin\textsuperscript{\rm 1,\rm 2},
    \equalcontrib Giacomo Lagomarsini\textsuperscript{\rm 1},
    Claudio Gallicchio\textsuperscript{\rm 3},
    Giuseppe Nuti\textsuperscript{\rm 1}
}
\affiliations {
    \textsuperscript{\rm 1}Silvretta Research, Inc.\\
    \textsuperscript{\rm 2}Department of Computer Science, Aalto University, Finland.\\
    \textsuperscript{\rm 3}Department of Computer Science, University of Pisa, Italy.\\
    andrea.perin@aalto.fi,
    g.lagomarsini@studenti.unipi.it,
    claudio.gallicchio@unipi.it,
    giuseppe.nuti@silvretta.ai
}

\nocopyright
\begin{document}
\maketitle
\begin{abstract}
We introduce a \textit{Mixture of Raytraced Experts}, a stacked Mixture of Experts (MoE) architecture which can dynamically select sequences of experts, producing computational graphs of variable width and depth.
Existing MoE architectures generally require a fixed amount of computation for a given sample. Our approach, in contrast, yields predictions with increasing accuracy as the computation cycles through the experts' sequence.
We train our model by iteratively sampling from a set of candidate experts, unfolding the sequence akin to how Recurrent Neural Networks are trained.
Our method does not require load-balancing mechanisms, and preliminary experiments show a reduction in training epochs of 10\% to 40\% with a comparable/higher accuracy.
These results point to new research directions in the field of MoEs, allowing the design of potentially faster and more expressive models.
The code is available at \href{https://github.com/nutig/RayTracing}{\texttt{nutig/RayTracing}}.
\end{abstract}


\section{Introduction}
MoEs have recently emerged as a popular construct in deep learning \citep[see]{ 10937907, mu2025}, particularly in language modeling, where the top-$k$ flavor is widely used.
Compared to a dense layer, these MoE models increase the number of trainable parameters whilst keeping the computational budget constant.
This is achieved by conditionally activating a small number $k$ of experts for a given sample, making top-$k$ MoEs examples of conditional computing \citep[see][for recent surveys]{9560049, Scardapane_2024}.

Here, we aim to address three additional challenges in conditional computing: proportionality of the resources used with the difficulty of the task; faster training by creating experts with homogeneous derivatives, without the need for load-balancing techniques; and availability of a solution, even if approximate, throughout the calculations' cycle, not just at the end of the process.

\begin{figure}
    \centering
    \includegraphics[width=\linewidth]{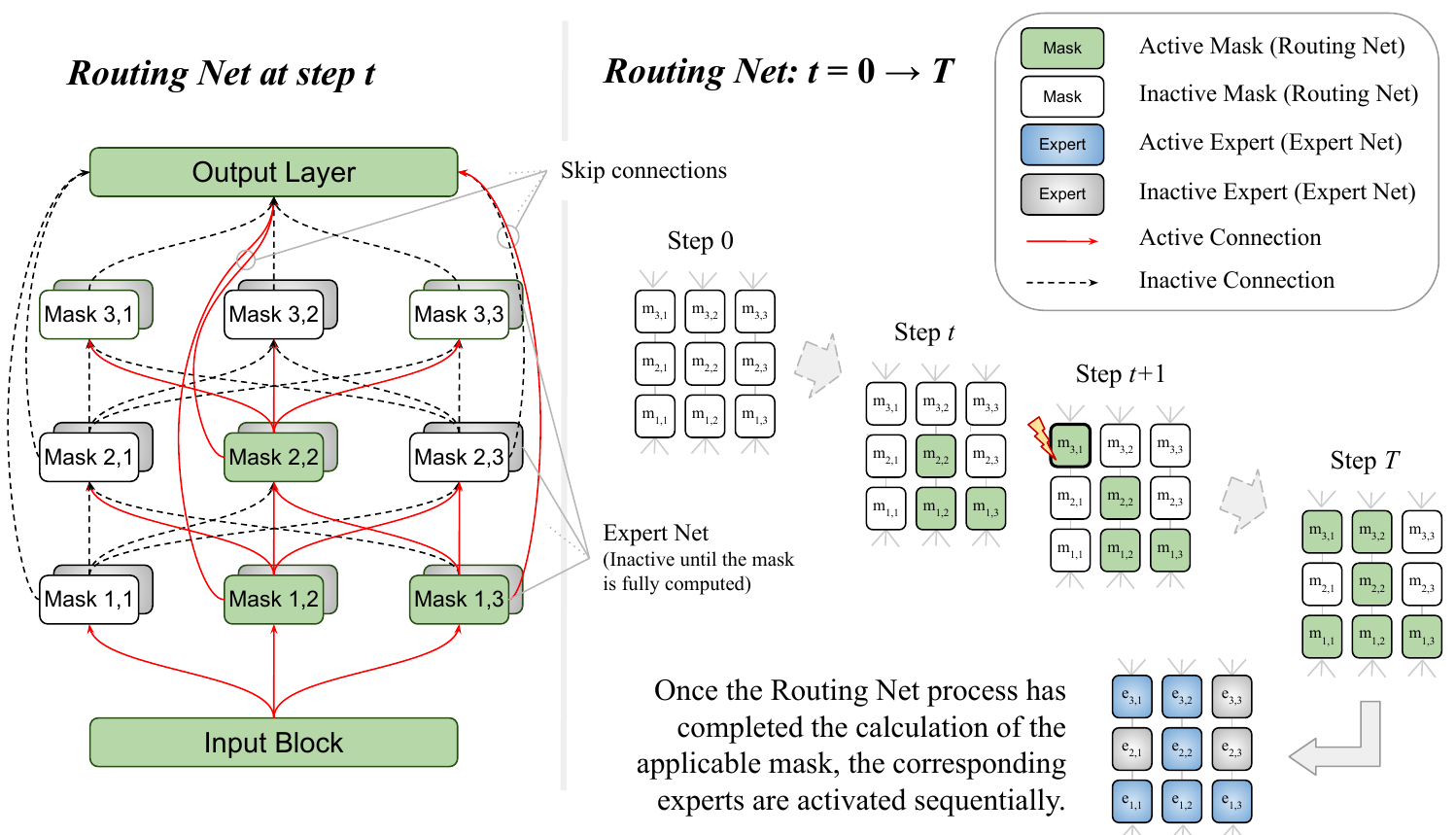}
    \caption{An iterative stochastic process determines the sequence of activated experts by computing an experts' mask: at each step, the next activated node is drawn from the eligible nodes proportionally to their firing rate. The probability of the next expert to be activated, among the eligible yet-to-be-activated experts -- in this example ($m_{1,1}$, $m_{2,1}$, $m_{2,3}$, $m_{3,2}$) -- is determined by the total incoming firing rates.
}
    \label{fig:RayTracing}
\end{figure}

Our core proposal is the architecture in Fig.~\ref{fig:RayTracing}. The model is centered on a routing network composed of stacked gates that control the experts' activation mask: an initial \emph{firing rate} is channeled through the routing network, iteratively activating the next node with a probability that is proportional to the total firing rate currently amassed in each yet-to-be-activated node. This allows us to search the sequence of experts that best fits an input -- training the sequence via unfolding of samples, akin to a recurrent neural network.

Theoretically, the routing network can be viewed as a set of nodes that receive activation rays (hence the model's name: \textit{Raytraced Experts}). A ray, originating via a Poisson process from the initial gate(s), is routed through the activated nodes -- with each node's softmax determining the probability of the ray's local route -- until it hits a yet-to-be-activated node (which becomes the next node in the activation sequence). By changing a ray's path probabilities to waiting time probabilities -- i.e. the time until a ray will activate each of the yet-to-be-activated nodes -- we are able to train the process efficiently. 

\begin{figure*}[ht]
    \centering
    \includegraphics[width=\linewidth]{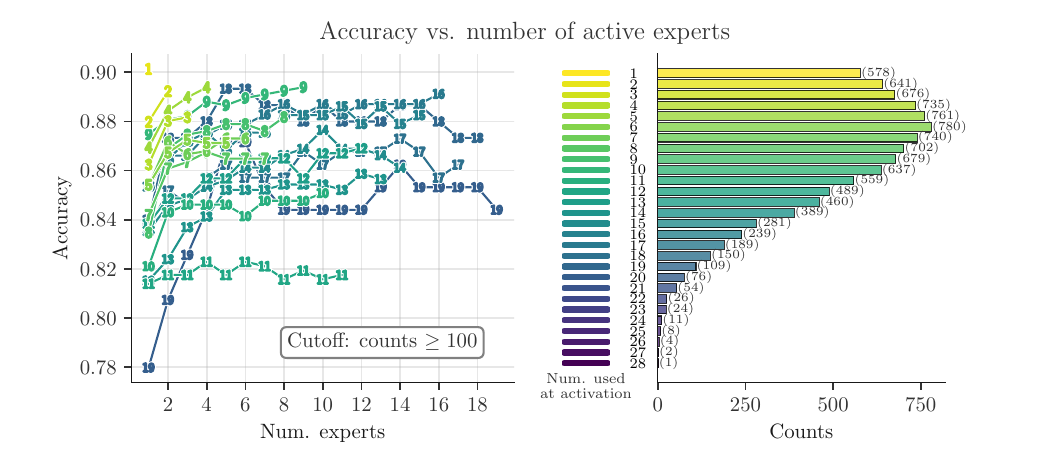}
\caption{\RT{} can dynamically adjust the number of experts based on sample difficulty.\
\textbf{Left:} Accuracy curves as the number of experts increases, up to the number actually selected by the model. We notice that samples using fewer experts achieve a higher accuracy. Moreover, using more experts increases accuracy on more difficult samples.\
\textbf{Right:} Histogram showing the distribution of expert counts used on the Fashion MNIST test set.}
    \label{fig:acc_curves}
\end{figure*}

\begin{figure*}
    \centering
    \includegraphics[width=\linewidth]{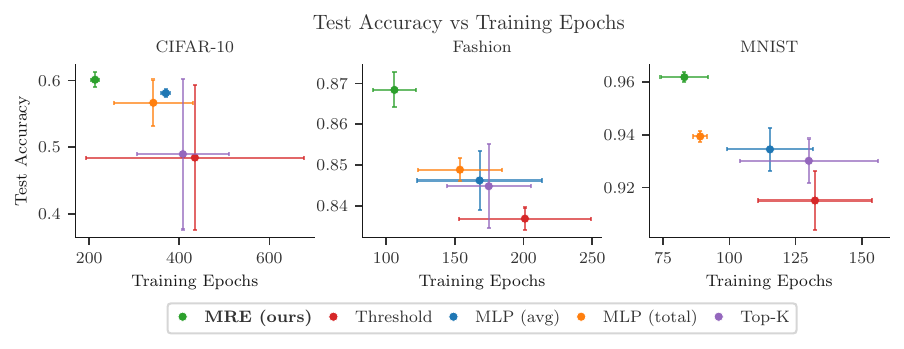}
    \caption{Test accuracy versus training epochs of \RT{} and baselines with equal learning rate set to \(10^{-3}\) (benchmark and baseline details are reported in Section \ref{sec:baseline}).}
    \label{fig:AccuracyEpochsScatter}
\end{figure*}

Assessing the results over three standard vision benchmarks (MNIST, Fashion-MNIST, and CIFAR), we can examine both the proportionality and sequential nature of the computation in the same chart (Fig.~\ref{fig:acc_curves}). The left panel displays the accuracy over the sequence of experts, with each line representing a different sequence length of a trained network. Importantly, this approach yields competitive accuracy with significantly fewer training epochs (Fig.~\ref{fig:AccuracyEpochsScatter}).

\section{Related Works}

MoE models were originally pioneered by \citet{shazeer2017outrageouslylargeneuralnetworks} and popularized by \citet{fedus2022switch}. More recently, it has become the \emph{de facto} standard for Large language Models (LLMs), \cite{10937907} and, at times, applied to vision \cite{han2024vimoe}. One of the more challenging aspect of MoEs revolves around load-balancing, which has dominated both academia and industry: recent studies present various solutions, such as leveraging input similarity across experts or expert-duplicating strategies \citep[see, e.g.,][]{omi2025loadbalancingmixtureexperts, feng2025comoecontrastiverepresentationmixtureofexperts, ma2025moegpsguidlinespredictionstrategy}. Interestingly, researchers at DeepSeek in \citet{wang2024auxiliarylossfreeloadbalancingstrategy} add a bias term to the gate (which is otherwise generally omitted) in order to facilitate a balanced use of experts -- but only on the forward pass, removing it during back-propagation so as not to pollute the gradients of the gates with an auxiliary function. A fundamental tension exists between the goal of load balancing and the objective of expert specialization, a concept thoroughly studied in \citet{guo2025advancingexpertspecializationbetter}.

Beyond standard routing mechanisms \citep[see][Sec. II, for a comprehensive taxonomy]{mu2025}, \citet{raposo2024mixtureofdepthsdynamicallyallocatingcompute} blend early-exit networks \citep{teerapittayanon2017branchynetfastinferenceearly} with an MoE construct. More recently, \citet{wang2025chainofexpertsunlockingcommunicationpower} iterate through multiple levels (of sometimes recursive and/or shared) experts, choosing the top-\emph{k} at every iteration. Both \citet{yue2025adak} and \citet{huang2024hardertasksneedexperts} present notable works on adaptive number of experts used. The former leverages Reinforcement Learning to determine the optimal set to use, whilst the latter employs a threshold for stopping a greedy cumulative gate probability mass expert(s) selection; we benchmark our approach to \citet{yue2025adak} and standard top-$k$ construct. The combination of dynamic expert choice, early-exit, and iterative expert selection is most similar with our proposed model.

Finally, on training: \citet{bengio2013estimating} originally researched stochastically activated networks, proposing an efficient yet biased approach. Since then, propagating derivatives through categorical choices is generally done via the Gumbel-softmax reparameterization trick \cite{jang2017categorical}, which is how we train this model. With a similar aim in mind as our approach, \citet{gu2025omniroutersharingroutingdecisions} propose sharing the same router across multiple layers in a bid to specialize chains of experts.

Complete technical details for this model, alongside ancillary definitions and exhaustive references, can be found in \citet{Nuti_2023, Nuti_2024a, Nuti_2024b, Nuti_2024c}. 

\section{Mixture of Raytraced Experts}
In our approach (in the following, \RT), for every sample, the network conditionally activates a subset of the available experts.
This subset is chosen by progressively including those as-of-yet unused experts which have the highest activation probability, herein termed \emph{firing rate}.
The firing rate can easily be translated into the probability of the expert having been activated at some discrete time unit $t$. 
This way, the algorithm determines an increasing sequence of active experts: it starts with a single expert in the first layer, and 
eventually ends with all of them being used.
At any point in the sequence, this set of active experts is effectively a \textit{computational path} linking the input block to the output block.
Along the sequence, these paths progressively involve more and more experts, and their contributions are integrated and used to refine the prediction of the system.

This sample-dependent sequence of computational paths is computed by using a \textit{routing network}.
If the stack of experts is a generalization of a shallow MoE layer, the routing network is a generalization of the standard gating mechanism usually seen in MoEs.

In practice, every expert is associated to a node in the routing network, establishing a parallelism between the decision mechanism (the routing) and the information mechanism (the experts).
Each of these routing nodes acts as a softmax gate, distributing its firing rate to its output nodes.
The cumulative firing rate that a node receives is used to rank and choose which as-of-yet unused node to include in the computational path.

\subsection{Activation sequences}
A defining feature of our architecture is the selection and usage, for each presented sample, of a \textit{sequence of experts}, as opposed to existing approaches which instead define a \textit{set of experts}.
By taking a sequential approach, we can define increasingly large families of experts, guided by the intuition that to a larger subset of experts corresponds a more precise output.
By indexing these families, we then associate a notion of ``time'' to the model, with the prediction being progressively refined as time goes on, taking more experts into account.
We call this sequence of experts the \textit{activation sequence} associated to a given sample.
The basic intuition is that we keep a running set of active nodes for each time step.
At every iteration, we add a node to the running set, picking from a set of candidate nodes.
The candidate nodes are those which are connected to a node among the active ones.
To each candidate, we associate its incoming firing rate, which serves as an (unnormalized) probability of being the next node to be picked.
This induces a categorical distribution over the candidates.  We sample the new candidate via the Gumbel-softmax trick in order to have a differentiable operation.

Crucially, among the candidates we consider a node corresponding to the output block.
The probability of choosing the output block is proportional to its incoming total firing rate.
The stopping condition is met when the output node is chosen\footnote{We tried other stopping conditions, e.g. when the output node's firing rate simply exceeds that of any other candidate. We chose the current method for its conceptual consistency.}: the network is considered to be active, and thus the set of experts that will be used is the one immediately preceding in the sequence.

The construction of the activation sequence is outlined in Algorithm~\ref{algo:act_sequence}.
We stress that, while the output activation set $\mathcal{A}^{(t)}$ takes binary values in $\{0, 1\}$, its value in the backward pass is \textit{path-dependent} due to the properties of the straight-through estimator (\texttt{ste} in Algorithm~\ref{algo:act_sequence}) computed via the Gumbel-Softmax trick \citep{jang2017categorical}.
We explored various temperature levels, and found that the results are robust with respect to this parameter, with a slight preference towards higher, exploration-promoting values (in a range between 10 and 50).

Since the output of the model is computed for a specific activation set $\mathcal{A}^{(t)}$, and since its construction depends on the previous value $\mathcal{A}^{(t-1)}$, the backpropagation takes this temporal dependence into account.
As a result, \RT{} is based on a novel \textit{sequential} construct.

\begin{algorithm*}
    \caption{Activation sequence}\label{algo:act_sequence}
    \hspace*{\algorithmicindent} \textbf{Input:} firing rates at first layer $\mathbf{f}_0\in\Delta^{N_e}$, Gumbel-softmax temperature $\tau\in \R_+$\\
    \hspace*{\algorithmicindent} \textbf{Output:} set of nodes at activation $\mathcal{A}^{(t)} \in\{0, 1\}^{L*N_e+1}$
    \begin{algorithmic}[1]
    \State $t \gets 0$
    \State $\mathcal{A}^{(0)} \gets \mathbf{0}\in\{0, 1\}^{L*N_e+1}$ \Comment{Set of active nodes at time $t$}
    \State $\mathcal{P}^{(0)} \gets \mathtt{concat}(\mathbf{f}_0, \mathbf{0} \in \R^{(\ell-1)* N_e + 1})$ \Comment{Probabilities of firing next}
    \While {$\mathcal{A}^{(t)}[-1] = 0$} \Comment{Output node is not active} 
        \State $t \gets t+1$
        \State $\mathcal{C}^{(t)} \gets \mathtt{ste}(\mathcal{P}^{(t-1)}, \tau)$ \Comment{Gumbel-softmax sampling from candidates}
        \State $\mathcal{A}^{(t)} \gets \mathcal{A}^{(t-1)} + \mathcal{C}^{(t)}$  \Comment{Add chosen candidate to active set}
        \State $\Tilde{\mathcal{A}}^{(t)} \gets \mathcal{A}^{(t)}[:-1]$ \Comment{Active set (excluding output node)}
        \State $\mathbf{f}^{(t)}, f_{out}^{(t)} \gets \mathtt{routing}(\mathbf{f}_0, \Tilde{\mathcal{A}}^{(t)})$ \Comment{Recompute firing rates for new active set}
        \State $\mathcal{P}^{(t)} \gets \mathtt{concat}(\mathbf{f}^{(t)} * (1-\Tilde{\mathcal{A}}^{(t)}), f_{out}^{(t)})$ \Comment{New candidate (unnormalized) probabilities}
    \EndWhile
    \end{algorithmic}
\end{algorithm*}

\subsection{Routing network}
\begin{figure}[ht]
    \centering
    \includegraphics[width=\linewidth]{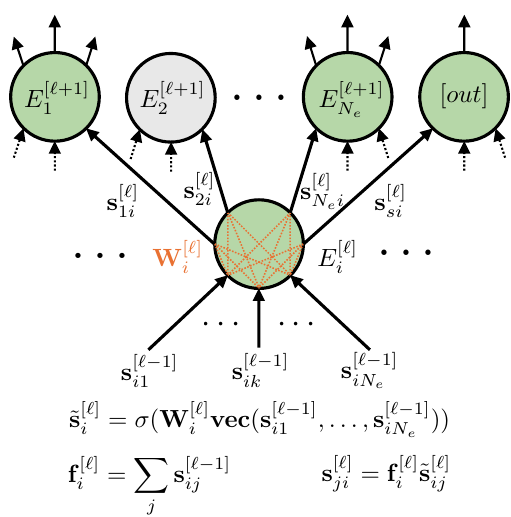}
    \caption{
    A pictorial representation of the routing mechanism for node $i$ at layer $\ell$.
    Inactive nodes (such as node 2 at layer $\ell+1$ in the picture) do not propagate their firing rate.
    The weight matrices $\mathbf{W}^{[\ell]}_i$, one for each node in the system, contain the trainable parameters of the routing mechanism.
    Each layer in the system can be easily implemented in parallel.
    }
    \label{fig:routing}
\end{figure}

The routing network is based on the idea of \textit{stacked and parallel} softmax gates.
To each expert we associate a gating node, so that the topologies of experts and gates mirror each other.
We assume in the following a rectangular grid with $L$ layers and $N_e$ nodes per layer. For any integer \(n>1\),
let \(\Delta^n \subset \R^n\) be the \(n-\)dimensional simplex.
For an input firing rate $\mathbf{f}^{(0)}\in\Delta^{N_e}$ and a set of active nodes $\mathcal{A}^{(t)}\in\{0,1\}^{L\times N_e}$, the routing network produces 1) a vector of \textit{firing rates} $\mathbf{f}^{(t)}\in\R_+^{L*N_e}$, one per expert, and 2) an output firing rate $f^{(t)}_{out}\in\R_+$.
We give an outline of the routing mechanism in Fig.~\ref{fig:routing}.

For node $i$ at layer $\ell$, we proceed as follows.
As a first step, the node receives its firing rate $\mathbf{f}^{[\ell]}_i$, computed as the sum of its inputs $\mathbf{s}^{[\ell-1]}_{ik}, k=1,\dots,N_e$.
To produce its outputs, a node first computes the weights $\Tilde{\mathbf{s}}^{[\ell]}_{ki}, k=1,\dots,N_e+1$ by mixing the inputs via the matrix $\mathbf{W}^{[\ell]}_i$ and then taking a softmax.
The firing rate of the node is then partitioned according to these weights, and the outputs are \(\mathbf{s}^{[\ell]}_i=\mathbf{f}^{[\ell]}_i\Tilde{\mathbf{s}}^{[\ell]}_i\).
Each node, in addition to the $N_e$ nodes in the next layer, is also connected to the output node ($[out]$ in the Fig.~\ref{fig:routing}) by \textit{skip connections}.
The firing rate of the output node is the sum of the incoming skip connections at every layer.

Inactive nodes, i.e., those not in $\mathcal{A}^{(t)}$, do not propagate their firing rate.
This is obtained by zeroing out the outputs of the node in question.
As a consequence,  the fewer active nodes, the smaller the total firing rate flowing through the network. 

\subsection{Expert Network}
The expert network is a collection of $L$ layers containing $N_e$ models each.
In this work, we assume that all the models share their architecture. In particular, we choose each expert as a small MLP.

At layer $\ell$, an  expert $E^{[\ell]}_i$ receives as its input the sum of the outputs of the (active) experts at layer $\ell-1$, \(\mathbf{h}^{[\ell-1]}\).
Calling $\mathcal{A}_i^{[\ell]}$ the value of the activation mask for the node, the output of layer \(\ell\) is again the sum of the outputs of each active expert of the layer,
\begin{equation}
    \mathbf{h}^{[\ell]} = \sum_{i=1}^{N_e} E_i^{[\ell]}(\mathbf{h}^{[\ell -1]}) \cdot \mathcal{A}^{[\ell]}_i.
\end{equation}
Unlike the standard approach used in top-\(k\) MoEs, in \RT{} we do not need to weight the experts' outputs by their corresponding probabilities, but only by the mask \(\mathcal{A}\),  which is made differentiable through the Gumbel-softmax trick.
 
 \section{Experimental Results}
\subsection{Baselines}\label{sec:baseline}
Here, we quickly illustrate the baselines that we compared \RT{} against.
We chose a classic multi-layer top-\(k\) MoE, with \(k=2\). 

We also considered a threshold based MoE, that, for each sample \(i\), activates the first \(k_i\) experts in order of router strength, until a threshold \(\vartheta\), which we fixed to \(0.5\) is reached. 
The MoEs baseline models, as well as our \RT, are all comprised of 4 layers of 8 experts each. 

\begin{figure*}
    \centering
    \includegraphics[width=\linewidth]{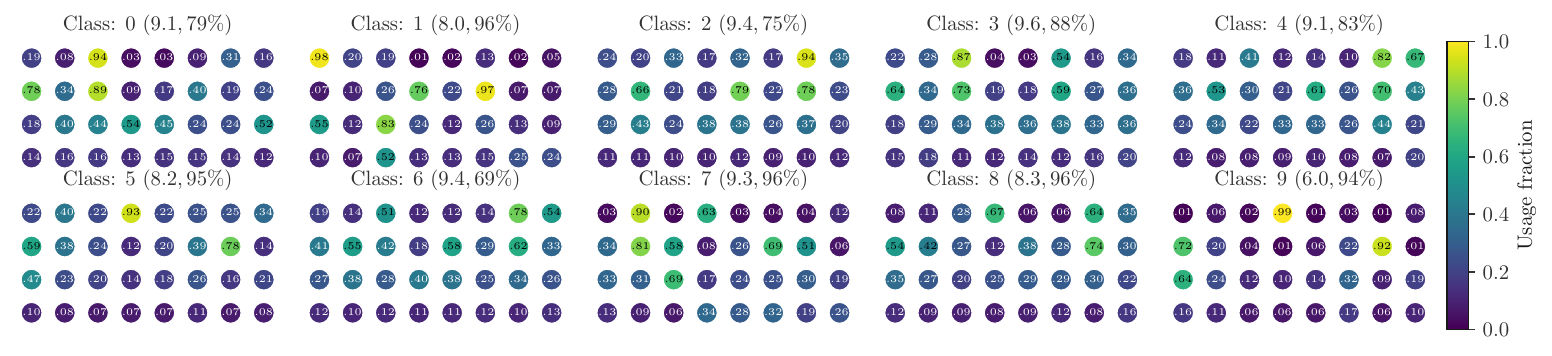}
    \caption{Samples of different classes activate different sets of experts (Fashion MNIST). Reported in parentheses are the average number of experts used and the accuracy rate for the class.
    Classes using fewer experts  (e.g. class 9 and class 1) are classified more accurately, while classes that use more experts (e.g. class 2 and 6) achieve lower accuracy score.}
    \label{fig:bitmap}
\end{figure*}

\begin{figure*}[ht]
    \centering
    \includegraphics[width=\linewidth]{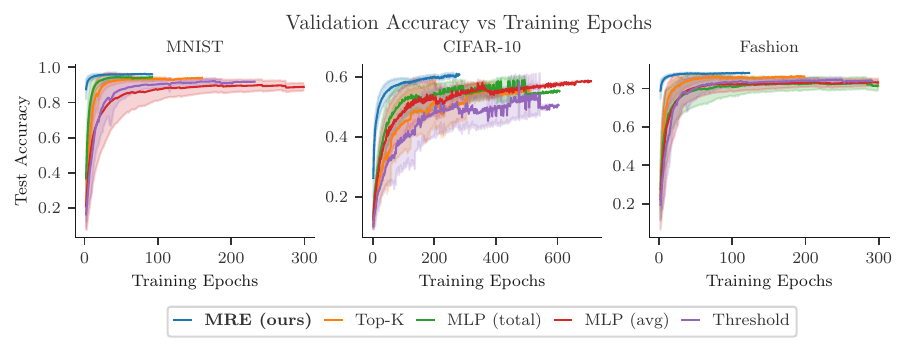}
    \caption{Validation accuracy over training epochs for the dataset considered. \RT{} reaches higher values in fewer epochs, compared to the baselines.}
    \label{fig:val-curves}
\end{figure*}

Finally, we also trained two MLPs, (named MLP total and MLP avg) having approximately the total number of parameters and the average number of parameters of our \RT{} models. All models, except MLP avg, have approximately \(40,000\) trainable parameters. 

\subsection{Results on Image Classification Tasks}
Table \ref{tab:results1} shows the results on 3 classic benchmark image classification tasks.
\RT{}  consistently outperforms other MoE architectures. 
Fig.~\ref{fig:bitmap} (Fashion MNIST)  shows how the routing network manages to route sample of different classes to different experts, while samples of similar classes are processed by similar experts (e.g. class 9 - ankle boots and class 5 - sandals).

Fig.~\ref{fig:val-curves} shows the validation accuracy over time.
On all datasets, our model uses a relatively small number of experts, that increases with the difficulty of the dataset (see Table \ref{tab:avg-params}). 

Fig.~\ref{fig:AccuracyEpochsScatter} shows the accuracy versus training epochs using the same learning rate of \(10^{-3}\). However, we note \RT{} seems to generally benefit from learning rates one order of magnitude larger than those of other MoE models, resulting in even  faster training. The optimal learning rate for \RT{} found is \(5\times 10^{-3}\), while for the baselines is \(5\times 10^{-4}\).
\begin{table}[h!]
\setlength{\tabcolsep}{4pt}
    \centering
    \begin{tabular}{lccccc}
    \toprule
         & MNIST & Fashion-MNIST & CIFAR10 &  \\
         \midrule
         MLP  (36) &95.3 (0.4)& 86.4 (0.1) & 59.7 (0.6)\\
         MLP (24) & 94.8 (0.2) & 85.3 (0.8) & 59.1 (0.1)\\
        Top-$k$ MoE & 95.3 (0.3) & 85.1 (0.4) & 55.7 (0.9) \\
        Threshold MoE & 93.1 (0.9) & 85.1 (0.2)  & 51.3 (6.1) \\
        \midrule
         \RT & 96.3 (0.1)  & 87.1 (0.1)  & 60.4 (1.2) \\
         \bottomrule
    \end{tabular}
    \caption{Results on classification tasks, averaged over 3 random initializations. Standard deviation reported in parentheses.}
    \label{tab:results1}
\end{table}

\begin{figure*}
    \centering
    \includegraphics[width=\linewidth]{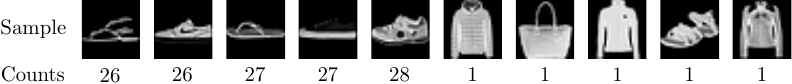}
    \caption{Example of difficulty analysis for a trained model on Fashion-MNIST. All the samples in this figure are correctly classified. We report the number of experts that have been used for each of the samples. A clear clustering in terms of used resources emerges.}
    \label{fig:diffanal}
\end{figure*}

\begin{table}[ht]
    \setlength{\tabcolsep}{4pt}
    \centering
    \begin{tabular}{cccc} 
        \toprule
         & Total params  &  avg params  & avg experts\\
         \midrule
         MNIST   & 40,202 & 20,275 & 7.5\\
         Fashion & 40,202 & 21,735 & 9.4  \\
         CIFAR10 & 33,330 & 16,520 & 11.4 \\
         \bottomrule
         & 
    \end{tabular}
    \caption{Total and average number of parameters for each dataset. Harder datasets use more experts, if the expert size is the same. CIFAR-10 has less total parameters because of the convolutional input layer, opposed to a fully-connected input layer for MNIST and Fashion.}
    \label{tab:avg-params}
\end{table}

\section{Conclusions}

We have developed an MoE model starting from a simple theoretical construct: a Poisson process firing calculation rays that activate a sequence of gates, corresponding to the experts' mask applicable to an input. The results show a balanced use of experts without needing a load-balancing mechanism. Strikingly, the process trains with fewer epochs than a comparable top-$k$ approach. We believe this may be due to the homogeneity of the derivatives within a \RT{} expert, driven by the correct formulation of the training process, which unfolds the experts sequencing as per a recurrent neural network --  though further research would shed more light as to the possible reason(s) for a shorter training cycle. Importantly, we would like to better understand how this approach can scale on larger datasets with experts that are truly onerous in compute time.
We will pursue both these research directions in the future.

\bibliography{references}

\appendix

\section{Routing network weight initialization}
We empirically observe that higher heterogeneity in the routing can be achieved by scaling the initialization weights $\mathbf{W}^{[\ell]}_i, \ell=1,\dots,L, i=1,\dots,N_e$ by $\sqrt{N}$, where $N$ is the number of output nodes.
For a routing network with skip connections, this amounts to scaling by $\sqrt{N_e+1}$.

To justify this choice, let us consider the following softmax operation, $y=\mathrm{softmax}(W x)$, where $y\in \R^N, x\in \R^G, W \in \R^{N\times G}$.
Further assume that we are at initialization, so that $W_{ij}\sim \mathcal{N}(0, \sigma^2)$ and $x\approx 1/N + \varepsilon u$, where $u\sim \mathcal{U}(\Delta^G)$.
We also assume $\varepsilon \ll 1/N$, which is reasonable when the number of nodes is sufficiently small.
Denoting the input to the softmax as $z=Wx$, we have
\begin{align*}
    \EX[z_j] = 0 \quad \mathrm{for\ all\ } j.
\end{align*}
Using the approximation for $x$, the variance is
\begin{align*}
 \Var[z_j] &\approx \Var[\sum_k W_{jk} (1/N + \varepsilon u)_k]\\
 &=(1/N)^2 \Var[\sum_k W_{jk}] + \varepsilon^2 \Var[\sum_k W_{jk}u_k].
\end{align*}
We ignore the $\varepsilon^2$ terms, obtaining
\begin{align*}
  \Var[z_j] \approx (1/N)^2 \sum_k W_{jk} = \sigma^2/N. 
\end{align*}
So, by choosing $\sigma^2=N$, we get approximately unit variance for $z$ at initialization.
Thus, successive routing layers effectively ``amplify'' the input signal (equivalently, the temperature is increased in the softmax).

%

\section{Architectural details}
\textbf{Input Block:}  For MNIST and Fashion MNIST, we use a single Fully connected linear layer to process the inputs, flattened into 784-dimensional vectors. 
For CIFAR-10, we use a very simple convolutional architecture, with
3 convolutional layers (4, 8 and 16 channels respectively) and 3 pooling layers interleaved.

\textbf{Experts:} For all the experiments in Table \ref{tab:results1}, we used a rectangular grid consisting of 4 layers of 8 experts each, for a total of 32 experts, we use fully connected 2-layers MLP experts, with each layer consisting of 16 neurons.

\textbf{MLPs:} The MLPs (total and avg) have a number of layer equal to the stacked MoE networks (8), and hidden sizes 36 and 24 respectively, roughly matching the total and effective number of parameters used by \RT.

\textbf{Output Layer:} For all the task, we use a fully connected linear layer to classify the samples.

\end{document}